\documentclass[acmsmall]{acmart}

\usepackage{fixltx2e}
\usepackage{amssymb}
\usepackage{booktabs}
\usepackage{multirow}
\usepackage{enumitem}

\def\BibTeX{{\rm B\kern-.05em{\sc i\kern-.025em b}\kern-.08emT\kern-.1667em\lower.7ex\hbox{E}\kern-.125emX}}
    
\copyrightyear{2018}
\acmYear{2018}
\setcopyright{acmlicensed}
\acmConference[Woodstock '18]{Woodstock '18: ACM Symposium on Neural Gaze Detection}{June 03--05, 2018}{Woodstock, NY}
\acmBooktitle{Woodstock '18: ACM Symposium on Neural Gaze Detection, June 03--05, 2018, Woodstock, NY}
\acmPrice{15.00}
\acmDOI{10.1145/1122445.1122456}
\acmISBN{978-1-4503-9999-9/18/06}

\begin{document}

%
\title{An Emotional Analysis of False Information in Social Media and News Articles}

%
\author{Bilal Ghanem}
\affiliation{%
  \institution{Universitat Polit\`ecnica de Val\`encia}
  \city{Valencia}
  \country{Spain}}
\email{bigha@doctor.upv.es}
\author{Paolo Rosso}
\affiliation{%
  \institution{Universitat Polit\`ecnica de Val\`encia}
  \city{Valencia}
  \country{Spain}}
\email{prosso@dsic.upv.es}
\author{Francisco Rangel}
\affiliation{%
  \institution{Universitat Polit\`ecnica de Val\`encia; Autoritas Consulting}
  \city{Valencia}
  \country{Spain}}
\email{francisco.rangel@autoritas.es}

%
\renewcommand\shortauthors{Ghanem, B. et al}

%
\begin{abstract}
Fake news is risky since it has been created to manipulate the readers' opinions and beliefs. In this work, we compared the language of false news to the real one of real news from an emotional perspective, considering a set of false information types (propaganda, hoax, clickbait, and satire) from social media and online news articles sources. Our experiments showed that false information has different emotional patterns in each of its types, and emotions play a key role in deceiving the reader. Based on that, we proposed a LSTM neural network model that is emotionally-infused to detect false news. \end{abstract}

%
%


%

%

%
\maketitle

\section{Introduction}
With the complicated political and economic situations in many countries, some agendas are publishing suspicious news to affect public opinions regarding specific issues \cite{nyhan2010corrections}. The spreading of this phenomenon is increasing recently with the large usage of social media and online news sources. Many anonymous accounts in social media platforms start to appear, as well as new online news agencies without presenting a clear identity of the owner. Twitter has recently detected a campaign\footnote{https://blog.twitter.com/official/en\_us/topics/company/2018/enabling-further-research-of-information-operations-on-twitter.html} organized by agencies from two different countries to affect the results of the last U.S. presidential elections of 2016. The initial disclosures by Twitter have included 3,841 accounts. A similar attempt was done by Facebook, as they detected coordinated efforts to influence U.S. politics ahead of the 2018 midterm elections\footnote{https://www.businessinsider.es/facebook-coordinated-effort-influence-2018-us-midterm-elections-2018-7}.

False information is categorized into 8 types\footnote{Types of False information: Fabricated, Propaganda,  Conspiracy Theories, Hoaxes, Biased or one-sided, Rumors, Clickbait, and Satire News.} according to \cite{zannettou2018web}. Some of these types are intentional to deceive where others are not. In this work, we are interested in analyzing 4 main types, i.e. \textbf{hoaxes}, \textbf{propagandas}, \textbf{clickbaits}, and \textbf{satires}. These types can be classified into two main categories - misinformation and disinformation - where misinformation considers false information that is published without the intent to deceive (e.g. satire). Disinformation can be seen as a specific kind of false information with the aim to mislead the reader (e.g. hoax, propaganda, and clickbait). 
\textbf{Propagandas} are fabricated stories spread to harm the interest of a particular party. \textbf{Hoaxes} are similar to propagandas but the main aim of the writer is not to manipulate the readers' opinions but to convince them of the validity of a paranoia-fueled story \cite{rashkin2017truth}. \textbf{Clickbait} is another type of disinformation that refers to the deliberate use of misleading  headlines, thumbnails, or stories' snippets to redirect attention (for traffic attention). \textbf{Satire} is the only type of misinformation, where the writer's main purpose is not to mislead the reader, but rather to deliver the story in an ironic way (to entertain or to be sarcastic).

The topic of fake news is gaining attention due to its risky consequences. A vast set of campaigns has been organized to tackle fake news. The owner of Wikipedia encyclopedia created the news site WikiTribune\footnote{https://www.wikitribune.com} to encourage the evidence-based journalism.

Another way of addressing this issue is by fact-checking websites. These websites like \textit{politifact.com}, \textit{snopes.com} and \textit{factchecking.org} aim to debunk false news by manually assess the credibility of claims that have been circulated massively in online platforms. These campaigns were not limited to the English language where other languages such as Arabic have been targeted by some sites like \textit{fatabyyano.net}\footnote{fatabyyano is an Arabic term which means ``to make sure''.}.

\paragraph{\textbf{Hypothesis}} Trusted news is recounting its content in a naturalistic way without attempting to affect the opinion of the reader. On the other hand, false news is taking advantage of the presented issue sensitivity to affect the readers' emotions which sequentially may affect their opinions as well.
A set of works has been done previously to investigate the language of false information. The authors in \cite{vosoughi2018spread} have studied rumours in Twitter. They have investigated a corpus of true and false tweets rumours from different aspects. From an emotional point of view, they found that false rumours inspired fear, disgust, and surprise in their replies while the true ones inspired joy and anticipation. Some kinds of false information are similar to other language phenomena. For example, satire by its definition showed similarity with irony language. The work in \cite{farias2016irony} showed that affective features work well in the detection of irony. In addition, they confirmed that positive words are more relevant for identifying sarcasm and negative words for irony \cite{wang2013irony}. The results of these works motivate us to investigate the impact of emotions on false news types. These are the research questions we aim to answer:

\begin{description}
\item \textbf{RQ1} \textit{Can emotional features help detecting false information?}
\item \textbf{RQ2} \textit{Do the emotions have similar importance distributions in both Twitter and news articles sources?}

\item \textbf{RQ3} \textit{Which of the emotions have a statistically significant difference between false information and truthful ones?}

\item \textbf{RQ4} \textit{What are the top-N emotions that discriminate false information types in both textual sources?}
\end{description}


In this work, we investigate suspicious news in two different sources: Twitter and online news articles.
Concerning the news articles source, we focus on the beginning part of them, since they are fairly long, and the emotional analysis could be biased by their length. We believe that the beginning part of false news articles can present a unique emotional pattern for each false information type since the writer in this part is normally trying to trigger some emotions in the reader.

Throughout the emotional analysis, we go beyond the superficial analysis of words. We hope that our findings in this work will contribute to fake news detection.

The key contributions of this article are:
\begin{itemize}
    \item \textbf{Model}:
We propose an approach that combines emotional information from documents in a deep neural network. We compare the obtained results with a set of baselines. The results show that our approach is promising.
    \item \textbf{Analysis}:
We show a comprehensive analysis on two false information datasets collected from social media and online news articles, based on a large set of emotions. We compare the differences from an affective perspective in both sources, and obtain valuable insights on how emotions can contribute to detect false news.
\end{itemize}

The rest of the paper is structured as follows; After a brief review of related work in Section~\ref{RW}, Section~\ref{EIN} introduces our emotionally-infused model. Then, we present the evaluation framework in Section~\ref{EF}. Section~\ref{ER} reports the experiments and the results, followed by an analysis on the false information types from emotional perspective in Section~\ref{D}. Finally, the conclusions of this work are summarized in Section~\ref{CF}.

\section{Related Work}
\label{RW}
The work that has been done previously on the analysis of false information is rather small regarding the approaches that were proposed. In this section, we present some recent works on the language analysis and detection of false information. 

Recent attempts tried to analyze the language of false news to give a better understanding. A work done in \cite{volkova2017separating} has studied the false information in Twitter from a linguistic perspective. The authors found that real tweets contain significantly fewer bias markers, hedges, subjective terms, and less harmful words. They also found that propaganda news targets morals more than satires and hoaxes but less than clickbaits. Furthermore, satirical news contains more loyalty and fewer betrayal morals compared to propaganda. In addition, they built a model that combined a set of features (graph-based, cues words, and syntax) and achieved a good performance comparing to other baselines (71\% vs. 59\% macro-F1). Another similar work \cite{rashkin2017truth} has been done to characterize the language of false information (propaganda, hoax, and satire) in online news articles. The authors have studied the language from different perspectives: the existence of weak and strong subjectivity, hedges, and the degree of dramatization using a lexicon from Wiktionary. As well, they employed in their study the LIWC dictionary to exploit the existence of personal pronouns, swear, sexual, etc. words. The results showed that false news types tend to use first and second personal pronouns more than truthful news. Moreover, the results showed that false news generally uses words to exaggerate (subjectives, superlatives, and modal adverbs), and specifically, the satire type uses more adverbs. Hoax stories tend to use fewer superlatives and comparatives, and propagandas use relatively more assertive verbs. Moving away from these previous false information types, the work in \cite{vosoughi2018spread} has focused on analyzing rumours in Twitter (from factuality perspective: True or False). They analyzed about 126,000 rumours and found that falsehood widespread significantly further, faster, deeper, and more broadly than truth in many domains. In addition, they found that false rumours are more novel than truthful ones, which made people more likely to share them. From an emotional perspective, they found that false rumours triggered "fear", "disgust", and "surprise" in replies while truthful ones triggered "anticipation", "sadness", "joy", and "trust". Another work \cite{kumar2016disinformation} has studied the problem of detecting hoaxes by analyzing features related to the content in Wikipedia. The work showed that some features like hoaxes articles' length as well as the ratio of wiki markups (images, references, links to other articles and to external URLs, etc.) are important to discriminate hoaxes from legitimate articles.
Many approaches have been proposed on fake news detection. In general, they are divided into social media and news claims-based approaches. The authors in \cite{qazvinian2011rumor, zhao2015enquiring, ruchansky2017csi, ma2016detecting, kochkina2018all} have proposed supervised methods using recurrent neural networks or by extracting manual features like a set of regular expressions, content-based, network-based etc. As an example, the work by \cite{castillo2011information} assessed the credibility of tweets by analyzing trending topics. They used message-based, user-based, and propagation-based features, and they found that some features related to the user information like user's age, number of followers, statuse counts etc. have helped the most to discriminate truthful from deceitful tweets. Other news claims-based approaches \cite{ghanem2018External, popat2016credibility, karadzhov2017fully, li2011t, popat2018declare} have been mainly focusing on inferring the credibility of the claims by retrieving evidences from Google or Bing search engines. These approaches have employed a different set of features starting from manual features (e.g. cosine similarity between the claims and the results, Alexa Rank of the evidence source, etc.) to a fully automatic approach using deep learning networks. A recent trend started to appear and is trying to approach the detection of fake news from a stance perspective. The aim is to predict how other articles orient to a specific fact \cite{ghanem2018stance, Hanselowski2018, Gaurav2018Combining}.

\section{Emotionally-infused Model}
\label{EIN}
In this section we describe the Emotionally-Infused Network we propose (EIN).

\subsection{Emotional Lexicons}
Several emotional models well-grounded in psychology science have been proposed, such as the ones by Magda Arnold \cite{arnold1960emotion}, Paul Ekman \cite{ekman1992argument}, Robert Plutchik \cite{plutchik2001nature}, and Gerrod Parrot \cite{parrott2001emotions}. On the basis of each of them, many emotional resources (lexicons) were built in the literature. In this work, we consider several emotional resources to increase the coverage of the emotional words in texts as well to have a wider range of emotions in the analysis. Concretely, we use EmoSenticNet, EmoLex, SentiSense, LIWC and Empath:

\begin{figure}[!htb]
  \includegraphics[width=9cm]{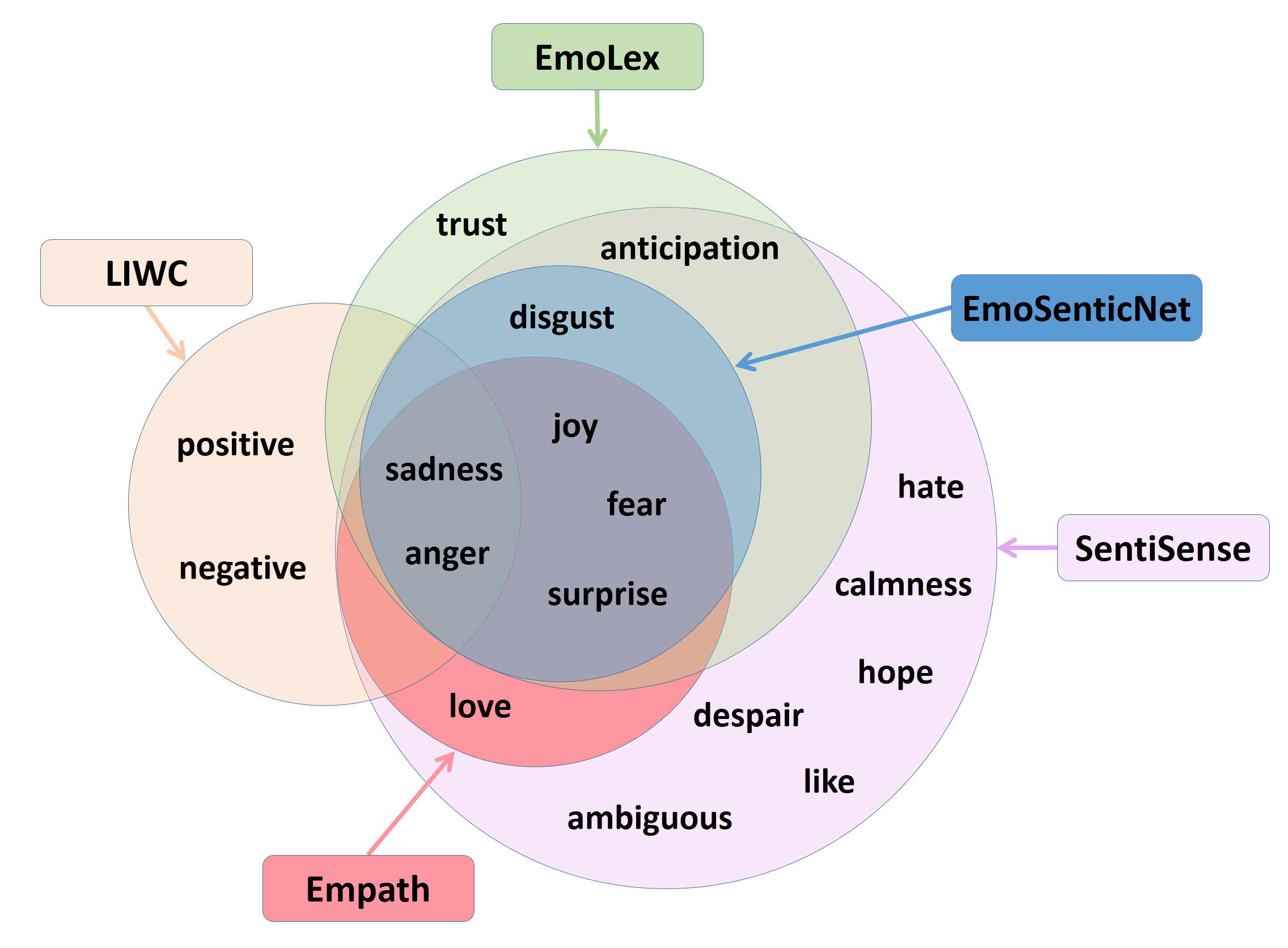}
  \caption{The emotional lexicons with their own emotions.}
  \label{fig:lexicons}
\end{figure}

\begin{itemize}
\item EmoSenticNet \cite{poria2013enhanced} is a lexical resource that assigns WordNet-Affect\footnote{http://wndomains.fbk.eu/wnaffect.html} emotion labels to SenticNet\footnote{https://sentic.net/} concepts. It has a total of 13,189 entries annotated using the six Ekman's basic emotions.
\item EmoLex \cite{mohammad2010emotions} is a word-emotion association lexicon that is labeled using the eight Plutchik's emotions. This lexicon contains 14,181 words.
\item SentiSense \cite{de2012sentisense} is a concept-based affective lexicon that attaches emotional meanings to concepts from the WordNet\footnote{https://wordnet.princeton.edu} lexical database. SentiSense has 5,496 words labeled with emotions from a set of 14 emotional categories, which is an edited version of the merge between Arnold, Plutchik, and Parrott models.
\item LIWC \cite{tausczik2010psychological} is a linguistic dictionary that contains 4,500 words categorized to analyze psycholinguistic patterns in text. Linguistic Inquiry and Word Count (LIWC) has 4 emotional categories: "sadness", "anger", "positive emotion", and "negative emotion".
\item Empath \cite{fast2016empath} is a tool that uses deep learning and word embeddings to build a semantically meaningful lexicon for concepts. Empath uses Parrott's model for the emotional representation, but we use only the primary emotions (6 emotions) in the Pattrott's hierarchy ("love", "joy", "surprise", "anger", "sadness", "fear").
\end{itemize}
In our study we consider the 17 emotions that we shown in Figure \ref{fig:lexicons}.

\subsection{Model}
\begin{figure}[!htb]
  \includegraphics[width=8cm]{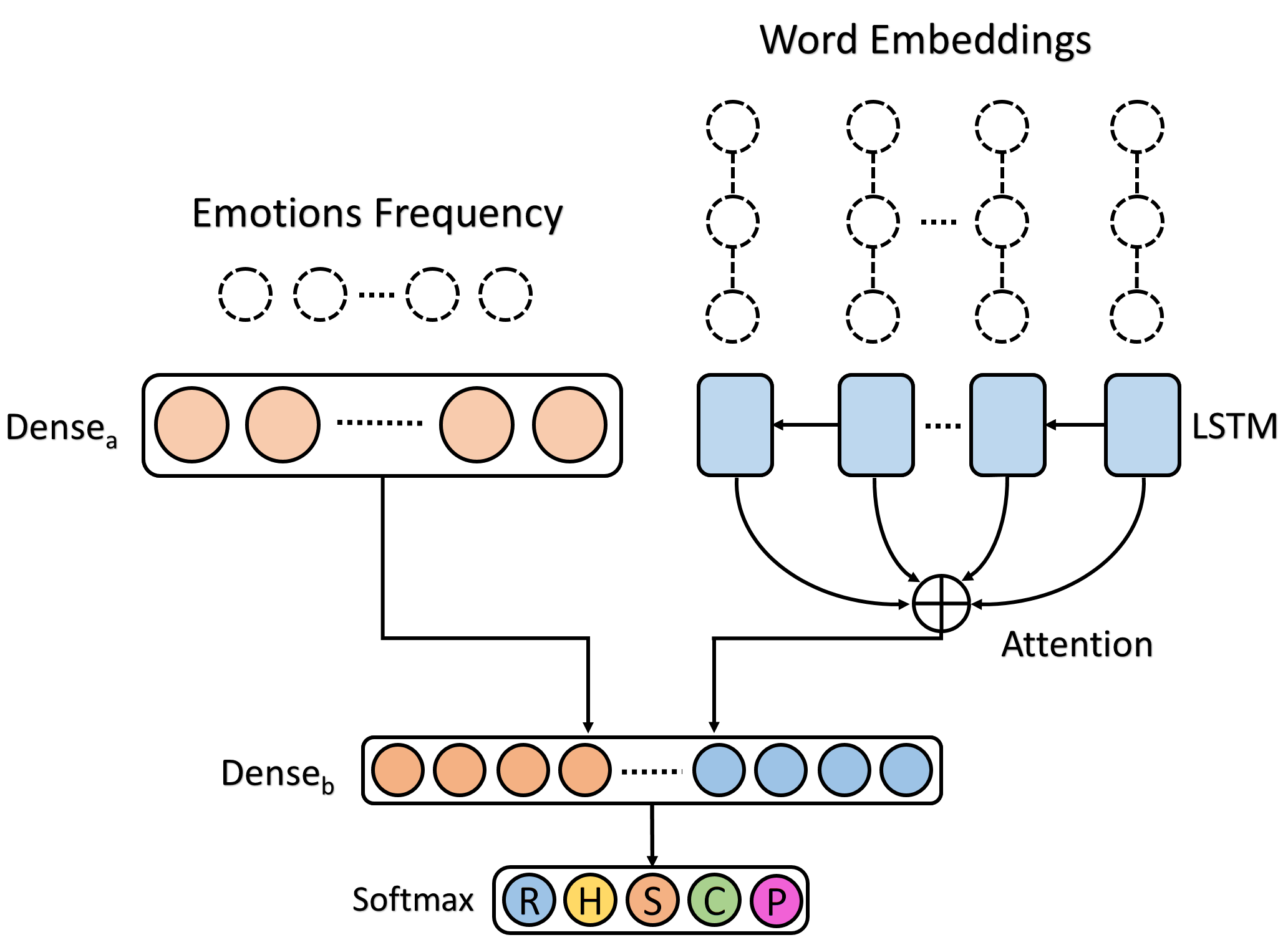}
  \caption{Emotionally-infused neural network architecture for false information detection.}
  \label{fig:architecture}
\end{figure}

We choose an Long short-term memory (LSTM) \cite{hochreiter1997long} that takes the sequence of words as input and predicts the false information type. The input of our network is based on word embedding (content-based) and emotional features (see Figure \ref{fig:architecture}).

\subsection{\textbf{Input Representation}}
Our network consists of two branches. In the content-based one, we use an embedding layer followed by a LSTM layer. Then, we add an attention layer \cite{raffel2015feed} to make this branch focus on (highlighting) particular words over others \footnote{We tested our model without the attention layer, but we got a lower result.}. The attention mechanism assigns a weight to each word vector result from the LSTM layer with a focus on the classification class. 
The input representation for this branch is represented as follows: the input sentence $S$ of length $n$ is represented as $[S\textsubscript{1}, S\textsubscript{2} .. S\textsubscript{n}]$ where $S\textsubscript{n} \in {\rm I\!R}^d$; ${\rm I\!R}^d$ is a d-dimensional word embedding vector of the $i$-th word in the input sentence. The output vectors of the words are passed to the LSTM layer, where the LSTM learns the hidden state $h\textsubscript{t}$ by capturing the previous timesteps (past features). The produced hidden state $h\textsubscript{t}$ at each time step is passed to the attention layer which computes a "context" vector $c\textsubscript{t}$ as the weighted mean of the state sequence $h$ by:

\begin{equation}
c\textsubscript{t} = \sum_{j=1}^{T} \alpha \textsubscript{tj}h\textsubscript{j},
\end{equation}

Where $T$ is the total number of timesteps in the input sequence and $\alpha \textsubscript{tj}$ is a weight computed at each time step
$j$ for each state h\textsubscript{j}. This output vector is then concatenated with the output from the dense\textsubscript{a} (see Figure \ref{fig:architecture}) layer and passed to the dense\textsubscript{b} layer, which precedes a final Softmax function to predict the output classes. Since the content-based branch is concatenated with the other emotional-based branch.\\
On the other hand, the input representation for the emotional-based branch is defined as follows: we have $N$ emotional lexicons $L\textsubscript{n}$ where $n\in[1, 5]$, each lexicon has $M$ number of emotions depending on the emotion model that the lexicon uses (e.g. Plutchik, Arnold, etc.). The emotion vector $E\textsubscript{m}$ of an input document using the $n$-th emotional lexicon is $L\textsubscript{n}E\textsubscript{m}$. In our implementation, the emotional vector $E\textsubscript{m}$ of a Lexicon $L\textsubscript{n}$ is built using word frequency and normalized by the input sentence's length. Each input sentence is represented using:
\begin{equation}
\label{eqn:emo_rep}
v=L\textsubscript{1}E\textsubscript{m} \oplus L\textsubscript{2}E\textsubscript{m} \oplus L\textsubscript{3}E\textsubscript{m} \oplus L\textsubscript{4}E\textsubscript{m} \oplus L\textsubscript{5}E\textsubscript{m},
\end{equation}

Where $v \in {\rm I\!R}^q$ and $q$ is:

\begin{equation}
\label{eqn:emo_rep_vec}
q = \sum_{i=1}^{n} || L\textsubscript{i}E\textsubscript{M} ||,
\end{equation}

\section{Evaluation Framework}
\label{EF}
\subsection{Datasets}
Annotated data is a crucial source of information to analyze false information. Current status of previous works lacks available datasets of false information, where the majority of the works focus on annotating datasets from a factuality perspective. However, to analyze the existence of emotions across different sources of news, we rely on two publicly available datasets and a list contains suspicious Twitter accounts.
\paragraph{\textbf{News Articles}} Our dataset source of news articles is described in \cite{rashkin2017truth}. This dataset was built from two different sources, for the trusted news (real news) they sampled news articles from the English Gigaword corpus. For the false news, they collected articles from seven different unreliable news sites. These news articles include satires, hoaxes, and propagandas but not clickbaits. Since we are interested also in analyzing clickbaits, we slice a sample from an available clickbait dataset \cite{chakraborty2016stop} that was originally collected from two sources: Wikinews articles' headlines and other online sites that are known to publish clickbaits. The satire, hoax, and propaganda news articles are considerably long (some of them reach the length of 5,000 words). This length could affect the quality of the analysis as we mentioned before. 
We focus on analyzing the initial part of the article. Our intuition is that it is where emotion-bearing words will be more frequent. 
Therefore, we shorten long news articles into a maximum length of N words (N=300). We choose the value of N based on the length of the shortest articles. Moreover, we process the dataset by removing very short articles, redundant articles or articles that do not have a textual content\footnote{e.g. "BUYING RATES US 31.170 .."}.

\paragraph{\textbf{Twitter}} For this dataset, we rely on a list of several Twitter accounts for each type of false information from \cite{volkova2017separating}. This list was created based on public resources that annotated suspicious Twitter accounts. The authors in \cite{volkova2017separating} have built a dataset by collecting tweets from these accounts and they made it available. For the real news, we merge this list with another 32 Twitter accounts from \cite{karduni2018can}. In this work we could not use the previous dataset\footnote{Due to Twitter terms of usage, the authors provided in their dataset the ids of the tweets and when we tried to collect these tweets many of them were deleted.} and we decide to collect tweets again. For each of these accounts, we collected the last M tweets posted (M=1000). By investigating these accounts manually, we found that many tweets just contain links without textual news. Therefore, to ensure of the quality of the crawled data, we chose a high value for M (also to have enough data). After the collecting process, we processed these tweets by removing duplicated, very short tweets, and tweets without textual content. 
Table \ref{tab:dataset} shows a summary for both datasets.

\begin{table}%
\caption{News articles and Twitter datasets' statistics.}
\label{tab:dataset}
\begin{minipage}{\columnwidth}
\begin{center}
\begin{tabular}{ccc}
  \toprule
  Category & News Articles & Twitter \\
  \hline
  Satire & 5,750 (18\%) & 12,502 (8\%) \\
  Hoax & 5,750 (18\%) & 6,247 (4\%) \\
  Propaganda & 5,750 (18\%) & 66,225 (43.5\%) \\
  Clickbait & 5,750 (18\%) & 36,103 (23.5\%)\\
  Real News & 8,550 (28\%) & 30,949 (21\%) \\
  \hline
  \textbf{Total} & \textbf{31,550} & \textbf{152,026} \\
  \bottomrule
\end{tabular}
\end{center}
\centering
\end{minipage}
\end{table}%

\subsection{\textbf{Baselines}} 
Emotions have been used in many natural language processing tasks and they showed their efficiency \cite{rangel2016impact}. We aim at investigating their efficiency to detect false information.
In addition to EIN, we created a model (Emotion-based Model) that uses emotional features only and compare it to two baselines. Our aim is to investigate if the emotional features independently can detect false news. The two baselines of this model are Majority Class baseline (MC) and the Random selection baseline (RAN). 

For the EIN model, we compare it to different baselines: \textbf{a}) The first one is bag-of-words with a support vector machine classifier (BOW-SVM). We test different classifiers, and we choose SVM since it gives the highest result in the 10-fold Cross Validation (CV); \textbf{b}) We use another baseline that is based on word embeddings where for each input document we extract an average word embedding vector by taking the mean of the embeddings for the document's words. Similarly, we test different classifiers and the Logistic Regression classifier shows the best performance (WE-LR); \textbf{c}) The last baseline is the same as our neural architecture but without the emotional features branch: an LSTM layer followed by attention and dense layers.

\section{Experiments and Results}
\label{ER}
\subsection{Emotion-based Model}
\label{can_emo}
In our experiments, we use $20\%$ of each of the datasets for testing and we apply $10$-fold cross-validation on the remain part for selecting the best classifier as well for tuning it. We tested many classifiers and we finally choose Random Forest for both datasets since it obtained the best results\footnote{The other classifiers that we tested are: Support vector machine (testing both kernels), naive bayes, logistic regression, k-nearest neighbor and multilayer perceptron.}. Table \ref{tab:rf} presents the classification results on both datasets.

\begin{table*}%
\centering
\caption{The results of the Emotion-based Model with the emotional features comparing to the baselines.}
\label{tab:rf}
\begin{minipage}{\columnwidth}
\begin{center}
\begin{tabular}{ccccc}
  \toprule
  { } & Accuracy & Macro-Precision & Macro-Recall & Macro-F1 \\
  \hline
  \multicolumn{5}{c}{\textbf{News Articles}}\\
  \hline
  Majority Class & 34.07 & 6.81 & 20.00 & 10.16 \\
  Random Selection & 20.55 & 20.66 & 20.87 & 20.20 \\
  Emotion-based Model & 49.80 & 47.57 & 50.68 & 48.15 \\
  \hline
  \multicolumn{5}{c}{\textbf{Twitter}}\\
  \hline
  Majority Class & 44.10 & 08.82 & 20.00 & 12.24 \\
  Random Selection & 20.40 & 20.33 & 20.35 & 17.83 \\
  Emotion-based Model & 51.71 & 54.95 & 37.86 & 41.38 \\
  \bottomrule
\end{tabular}
\end{center}
\centering
\end{minipage}
\end{table*}%

The results in both datasets show that emotional features clearly detect false news, compared to the baselines (\textbf{RQ1}). The emotional features perform better in the news articles dataset compared with these of tweets. We are interested in investigating also how good are the emotional features in detecting each class comparing to the RAN baseline. We choose the RAN baseline since it shows better results with regard to macro-F1 score. For doing so, we investigated the True Positive (TP) classification ratio for each class in each dataset. 


The clickbait class shows the highest TPs comparing to the other classes.  From this we can infer that clickbaits exploit emotions much more than the other classes to deceive the reader. It is worth to mention that for the hoax class the proposed approach is better than the random baselines with a small ratio ($4\%$ difference). This could be justified by the fact that hoaxes, by definition, try to convince the reader of the credibility of a false story. Hence, the writer tries to deliver the story in a normal way without allowing the reader to fall under suspicion.
The number of instances related to the false information classes in the news articles dataset is the same. Therefore, there is not a majority class that the classifier can be biased to. This is not the case in the Twitter dataset.
For the Twitter dataset, the dataset is not balanced. Therefore, where the results are biased by the majority class (propaganda). But in general, all the classes' TP ratios are larger than the corresponding ones obtained with RAN baseline.
From these results, we can conclude that suspicious news exploits emotions with the aim to mislead the reader. Following, we present the results obtained by the proposed emotionally-infused model.

\subsection{Emotionally-Infused Model}
In the neural model, to reduce the computational costs, instead of the cross-validation process we take another $20\%$ from the training part as a validation set\footnote{We are forced to use different validation scenarios because for selecting the best parameters in the classical machine learning Scikit-Learn library we used Grid Search technique where CV is the only option for tuning. On the other hand, it is too expensive computationally to use CV to tune a deep neural network using a large parameter space.} (other than the $20\%$ that is prepared for testing). For the pretrained word embeddings, we use Google News Word2Vec 300-Embeddings\footnote{https://code.google.com/archive/p/word2vec/} in the neural network as well as in the W2V-LR baseline. For the classical machine learning classifiers for the baselines, we use the Scikit-Learn python library, and for the deep learning network, we use Keras library with Tensorflow as backend. To tune our deep learning network (hyper-parameters), we use the Hyperopt\footnote{https://github.com/hyperopt/hyperopt} library. And to reduce the effect of overfitting, we use early stopping technique.

In Table \ref{tab:params} we summarize the parameters with respect to each dataset. We have to mention that we use Dropout after the dense layer in the emotional features branch (Drop\textsubscript{c}) as well as after the attention layer in the other one (Drop\textsubscript{d}) before the concatenation process. Since it is a multiclass classification process, we use categorical cross-entropy loss function. A summary of the models' parameters is presented in Table \ref{tab:params}.

\begin{table*}%
\centering
\caption{Models' parameters used in the three datasets (News articles, Twitter, Stop\_Clickbaits). LSTM: the 3rd baseline, EIN: Emotionally-Infused Network.}
\label{tab:params}
\begin{minipage}{\columnwidth}
\begin{center}
\begin{tabular}{ccccccc}
  \toprule
  
  \multirow{2}{*}{Parameter} & \multicolumn{2}{c}{News Articles} & \multicolumn{2}{c}{Twitter} & \multicolumn{2}{c}{Stop\_Clickbait} \\
   & LSTM & EIN & LSTM & EIN & LSTM & EIN \\
  \hline
  LSTM units                    & 140 & 90 & 180 & 180 & 120 & 120 \\
  Dense\textsubscript{a} units  & -   & 320 & - & 100 & - & 60 \\
  Dense\textsubscript{b} units  & 320 & 60 & 120 & 60 & 260 & 120 \\
  Batch                         & 64  & 64 & 64 & 64 & 32 & 32 \\
  Activation                    & relu  & relu & relu & relu & tanh & relu \\
  optimizer                     & adadelta & adam & adadelta & rmsprop & rmsprop & Adam \\
  Drop\textsubscript{c}         & 0.5 & 0.5 & 0.5 & 0.2 & 0.2 & 0.2 \\
  Drop\textsubscript{d}         & 0.2 & 0.1 & 0.2 & 0.2 & 0.2 & 0.2 \\
  \bottomrule
\end{tabular}
\end{center}
\centering
\end{minipage}
\end{table*}%

Table \ref{tab:EIN_results} summarizes the performance of the proposed model in comparison to those obtained by the baselines. We report Macro- precision, recall, and F1, including also the metric of accuracy; for comparing the models' results we consider the macro of metrics since it shows an averaged result over all the classes. The baselines that we propose clearly show high results, where the LSTM baseline has the best performance in news articles dataset. In Twitter there is a different scenario, the BOW-SVM baseline shows a higher performance with respect to LSTM. We are interested in investigating the reason behind that. Therefore, we checked the coverage ratio of the used embeddings in the Twitter dataset. We have to mention that we excluded stop words during representing the input documents using the pre-trained Google News word embeddings\footnote{The existence of stop words is importance to conserve the context in the LSTM network, but we got better results without them.}. In the news articles dataset, we found that the coverage ratio of the embeddings is around $94\%$ while in Twitter it is around $70\%$. Therefore, we tuned the word embeddings during the training process to improve the document's representation since we have a larger dataset from Twitter. This process contributed with $1.9\%$ on the final macro-F1 results in Twitter (the result without tuning is $53.51\%$). Even though, the results obtained with the LSTM baseline is still lower than the one obtained with BOW-SVM. This experiment gives us some intuition that the weaker performance on Twitter may be due to the embeddings. Therefore, we tried different embeddings but none of them improved the result\footnote{e.g. Glove (using multiple embedding dimensions) and FastText.}. The second baseline (W2V-LR) proved the same issue regarding the embeddings. The W2V-LR macro-F1 result in the news articles dataset is competitive, where it is much lower in Twitter. The usage of LSTM is two folds: in addition to being a good baseline, it shows also how much the emotional features contribute in the emotionally-infused network. 

EIN results outperform the baselines with a large margin (around 2\% in Twitter and 7\% in news articles), especially in the news articles dataset. The margin between EIN and the best baseline is lower in the Twitter dataset. The results also show that combining emotional features clearly boosts the performance. We can figure out the improvement by comparing the results of EIN to LSTM. EIN shows superior results in news articles dataset with regard to the LSTM (79.43\%). A similar case appears in the Twitter dataset but with a lower margin (59.70\%). The results of EIN in Twitter dataset show that emotional features help the weak coverage of word embeddings to improve the performance as well as to overcome the BOW-SVM baseline.

\begin{table*}%
\centering
\caption{Results of the proposed model (EIN) vs. the baselines.}
\label{tab:EIN_results}
\begin{minipage}{\columnwidth}
\begin{center}
\begin{tabular}{ccccc}
  \toprule
  { } & Accuracy & Macro-Precision & Macro-Recall & Macro-F1 \\
  \hline
  \multicolumn{5}{c}{\textbf{News Articles}}\\
  \hline
  BOW+SVM & 73.67 & 71.81 & 71.11 & 70.70 \\
  W2V+LR & 72.11 & 69.97 & 70.28 & 69.78 \\
  LSTM & 74.79 & 79.69 & 70.85 & 72.26 \\
  EIN & 80.72 & 79.52 & 79.82 & 79.43 \\
  \hline
  \multicolumn{5}{c}{\textbf{Twitter}}\\
  \hline
  BOW+SVM & 62.86 & 59.53 & 55.94 & 57.45 \\
  W2V+LR & 52.71 & 48.58 & 35.10 & 36.43 \\
  LSTM & 63.29 & 64.44 & 52.00 & 55.41 \\
  EIN & 64.82 & 60.65 & 58.90 & 59.70 \\
  \bottomrule
\end{tabular}
\end{center}
\centering
\end{minipage}
\end{table*}%

\begin{figure*}[!htb]
\minipage{0.5\textwidth}
\centering
  \includegraphics[width=6.5cm]{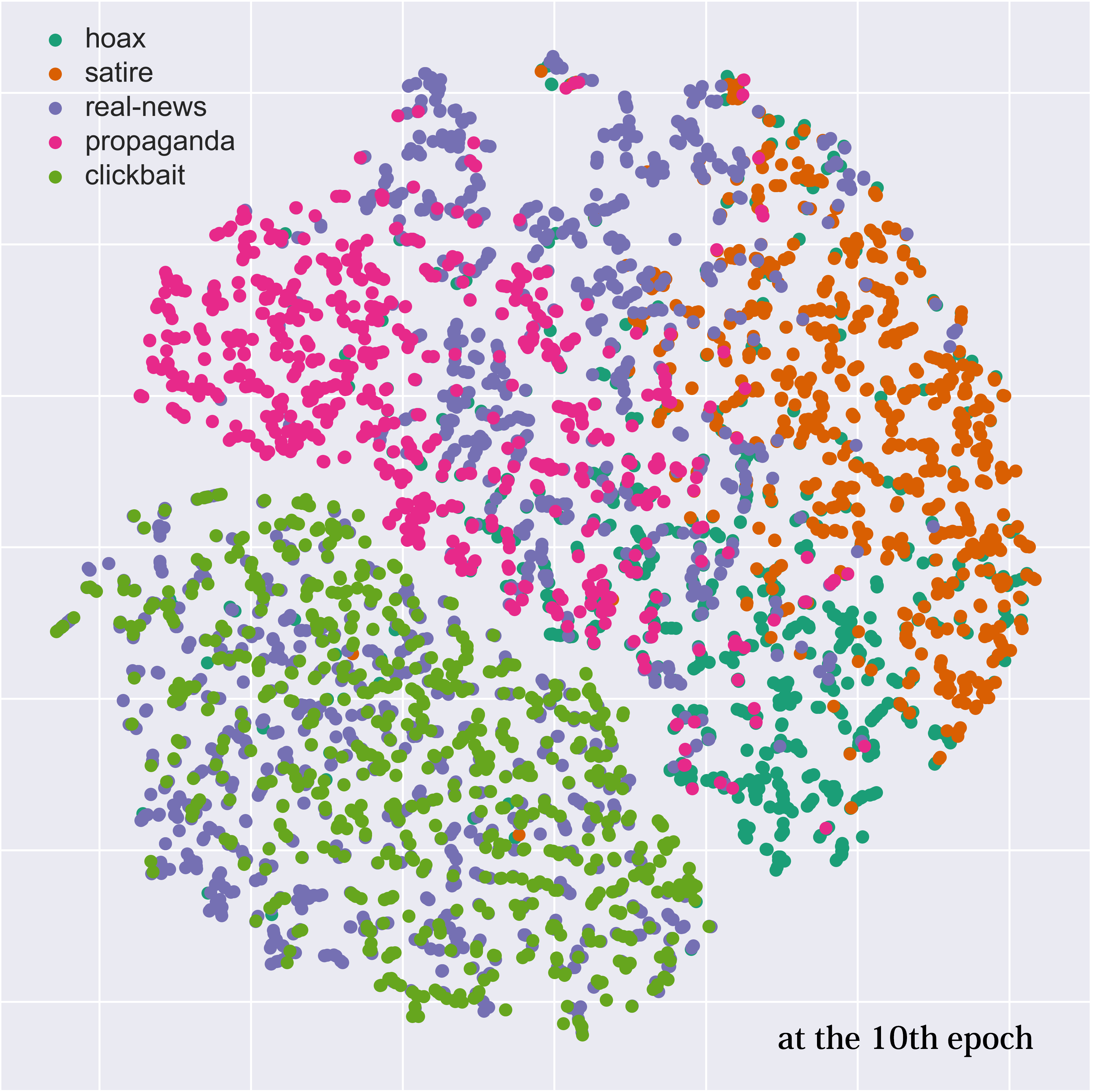}
\endminipage\hfill
\minipage{0.5\textwidth}%
\centering
  \includegraphics[width=6.5cm]{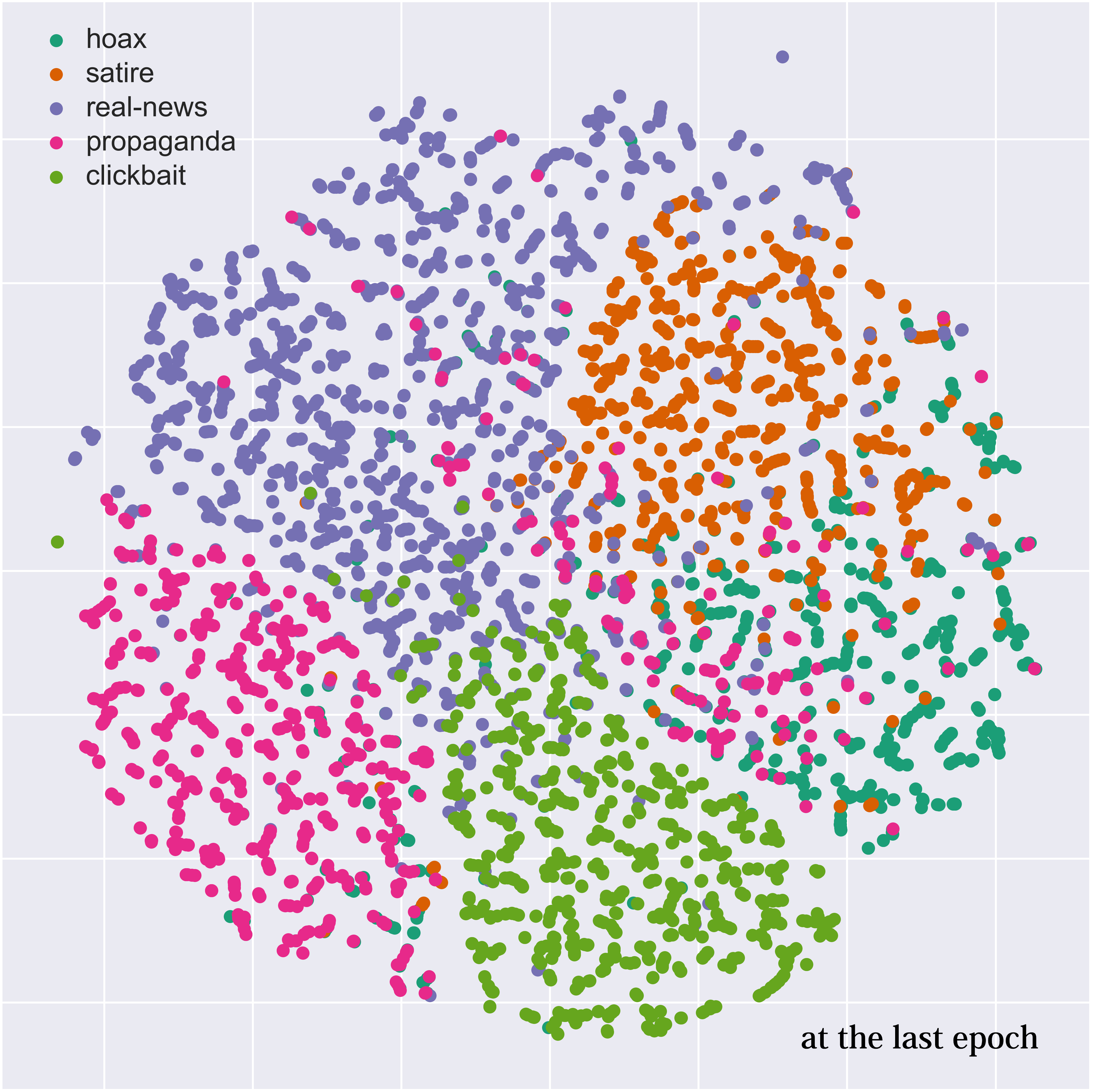}
\endminipage
  \caption{Projection of documents representation from the news articles dataset.}
  \label{fig:tsne}
\end{figure*}

We observed before that clickbait TP's ratio of the news articles dataset is the highest one, and this result points out that the clickbait class is less difficult to detect specifically from an emotional perspective. Therefore, in order to assess how our model separates false information types, we employ dimensionality reduction using t-distributed Stochastic Neighbor Embedding (T-SNE) technique \cite{maaten2008visualizing} to project the document's representation from a high dimensional space to a 2D plane. Thus, we project the embeddings in EIN by extracting them from the outputs of Dense\textsubscript{b} layer (see Figure \ref{fig:tsne}). We extract the embeddings twice, once from a random epoch (epoch 10) at the beginning of the training phase and the other at the last epoch. 

Our aim from the early epoch projection is to validate what we have noticed: the clickbait class is less difficult to detect with regard to the other classes.
As we can notice in the 10-epoch plot, the clickbait class needs few epochs to be separated from the other types, and this supports what we found previously in the manual investigation of the classes' TP ratios. Despite this clear separation, there is still an overlapping with some real-news records. This results points out that emotions in clickbaits play a key role in deceiving the reader. Also, the figure shows that the disinformation classes still need more training epochs for better separation. Real-news records are totally overlapped with the false information classes as well as the false information classes with each other. 
On the other hand, for the last epoch, clearly, the classes are separated from each other and the more important, from the real news. But generally, there still a small overlapping between satires and hoaxes as well few records from the propaganda class.

\subsection{EIN as Clickbaits Detector}
From the previous results in Section \ref{can_emo} as well as from what we notice in Figure \ref{fig:tsne}, EIN obtains a clear separability of the clickbait class. These observations motivate us to investigate EIN as clickbait detector. Concretely, we test EIN on the source of our clickbait instances \cite{chakraborty2016stop} in the news articles dataset. As we mentioned previously, this dataset originally was built using two different text sources. For clickbaits, the authors have manually identified a set of online sites that publish many clickbait articles. Whereas for the negative class, they collected headlines from a corpus of Wikinews articles collected in other research work. They took 7,500 samples from each class for the final version of the dataset. The authors also proposed a clickbaits detector model (Stop\_Clickbait) that employed a combination of features: \textit{sentence structure} (sentence length, average length of words, the ratio of the number of stop words to the number of thematic words and the longest separation between the syntactically dependent words), \textit{word patterns} (presence of cardinal number at the beginning of the sentence, presence of unusual punctuation patterns), \textit{clickbait language} (presence of hyperbolic words, common clickbait phrases, internet slangs and determiners), and \textit{N-grams features} (word, Part-Of-Speech, and syntactic n-grams). Using this set of features group, the authors tested different classifiers where SVM showed the state-of-the-art results. They considered Accuracy, Precision, Recall and F1 to compare their approach to a baseline (an online web browser extension for clickbaits detection called Downworthy\footnote{http://downworthy.snipe.net/}).

\begin{table}%
\caption{The performance of EIN on the clickbaits dataset using 10-fold CV.}
\label{tab:stop_CB}
\begin{minipage}{\columnwidth}
\begin{center}
\begin{tabular}{ccccc}
  \toprule
  { } & Accuracy & Precision & Recall & F1\\
  \hline
  Stop\_Clickbait & 93 & 95 & 90 & 93 \\
  LSTM & 95.89 & 95.58 & 96.26 & 95.91 \\
  EIN & 96.31 & 95.74 & 96.97 & 96.35 \\
  \hline
  \bottomrule
\end{tabular}
\end{center}
\centering
\end{minipage}
\end{table}%

In this experiment, we consider the third baseline (LSTM) to observe the improvement of the emotional features in the EIN model. Different from the previous experiments, this is a binary classification task. Therefore, we use binary cross-entropy as loss function and we change the Softmax layer to a Sigmoid function. The new parameters for both LSTM and EIN models are mentioned in Table \ref{tab:params}.

In Table \ref{tab:stop_CB} we present the results of the Stop\_Clickbait approach, LSTM baseline, and the EIN model. The results show that our baseline outperforms the proposed clickbait detector with a good margin. Furthermore, the results of the EIN are superior to the LSTM and the Stop\_Clickbait detector. Considering emotions in the EIN deep learning approach improved the detection of false information. This is due to the fact that in clickbaits emotions are employed to deceive the reader.

\section{Discussion}
\label{D}
The results show that the detection of suspicious news in Twitter is harder than detecting them in news articles. Overall, the results of EIN showed that emotional features improve the performance of our model, especially in the case of the news articles dataset. We manually inspected the Twitter dataset and observed that the language of the tweets has differences compared to the news articles one. We found that news in Twitter has many abbreviations (amp, wrt, JFK...etc.), bad words abbreviations (WTF, LMFO...etc.), informal language presentation, and typos. This reduces the coverage ratio of word embeddings. We also noticed that suspicious news in Twitter are more related to sexual issues. To validate our observations, we extracted the mean value of sexual words using a list of sexual terms \cite{frenda2018exploration}. The mean value is the average number of times a sexual/bad word appears in a tweet normalized by the length of the tweet. The mean value in Twitter is 0.003\footnote{The mean value is normalized by the sentence length since the news articles documents are longer than Tweets.} while in news articles is 0.0024. Similarly, suspicious news in Twitter presented more insulting words\footnote{Insult-wiki: http://www.insult.wiki/wiki/Insult\_List} than in news articles where the mean value in Twitter is 0.0027 and 0.0017 in news articles.

Following, we focus on analyzing false information from an emotional perspective. 
We are aiming to answer the rest of the questions, \textbf{RQ2}, \textbf{RQ3}, and \textbf{RQ4}.





\textbf{RQ2} \textit{Do the emotions have similar importance distributions in both Twitter and news articles sources?}

\begin{figure}[!htb]
  \includegraphics[width=8cm]{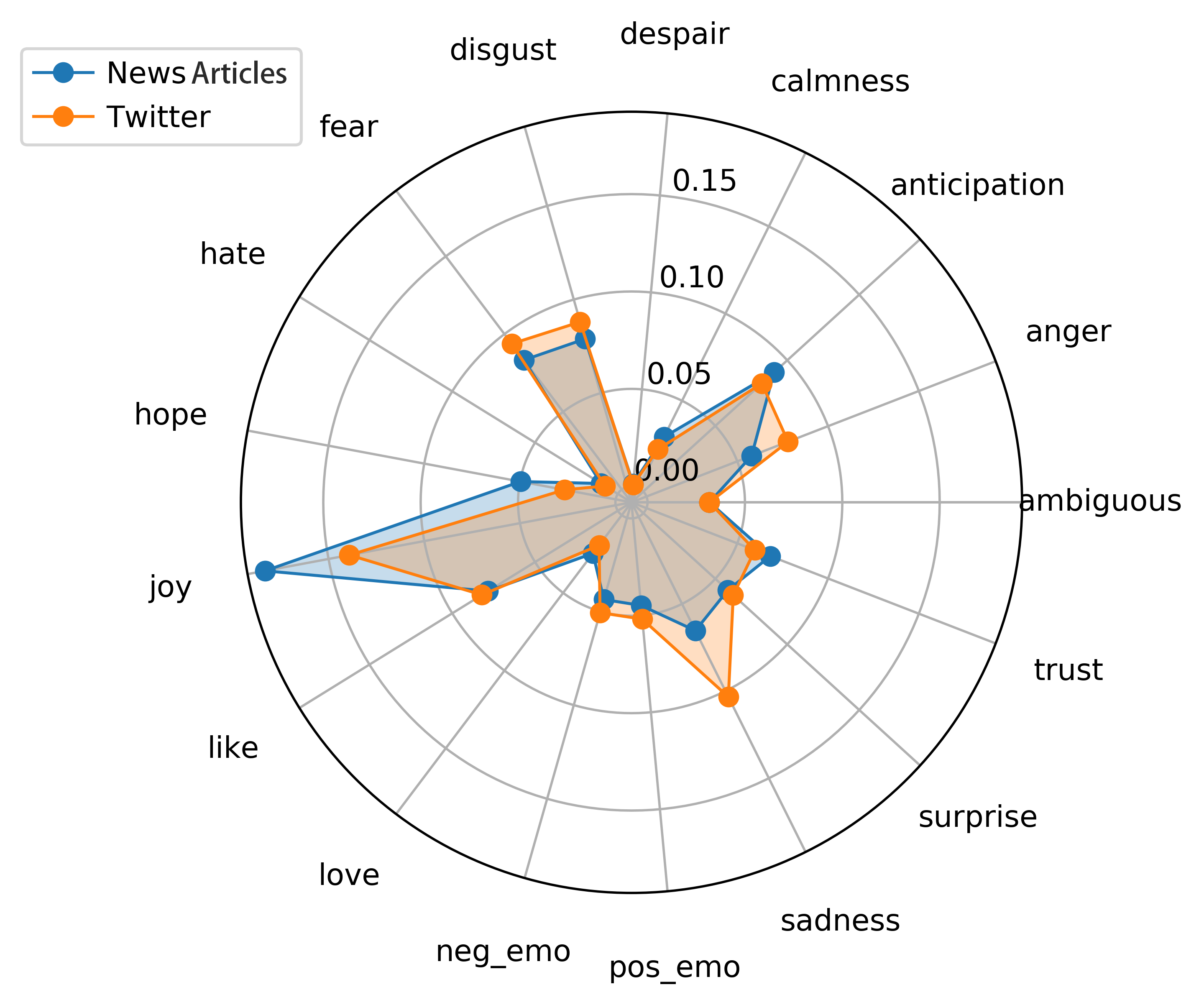}
  \caption{Best ranked features according to Information Gain.}
  \label{fig:cross_emotions}
\end{figure}

Intuitively, the emotions contribution in the classification process is not the same, where some words could manifest the existence of specific kind of emotions rather than others. To investigate this point, we use Information Gain (IG) in order to identify the importance of emotions in discriminating between real and all the other types of false news (multiclass task) in both Twitter and news articles datasets (see Figure \ref{fig:cross_emotions}). Before going through the ranking of features importance, we notice that the emotions ranking shapes are very similar in both Twitter and news articles. This states that despite the fact that the language is different, both sources have similar overall emotions distribution. In other words, false news employs a similar emotional pattern in both text sources.
Since the news language in Twitter is not presented clearly as in news articles, this observation can help to build a cross-source system that is trained on suspicious news from news articles to detect the corresponding ones in Twitter. Figure \ref{fig:cross_emotions} shows also that the emotion "joy" is the most important emotion in both datasets. It also mentions that "despair" and "hate" are almost not used in the classification process. The ranking of the features in both sources is different, where in the news articles dataset the top important emotions are "joy", "anticipation", "fear", and "disgust" respectively. On the other hand, the top ones in Twitter are "joy", "sadness", "fear", and "disgust".

{ \color{white}.}

\textbf{RQ3} \textit{Which of the emotions have a statistically significant difference between false information and truthful ones?}

\begin{figure}[!htb]
  \includegraphics[width=4cm]{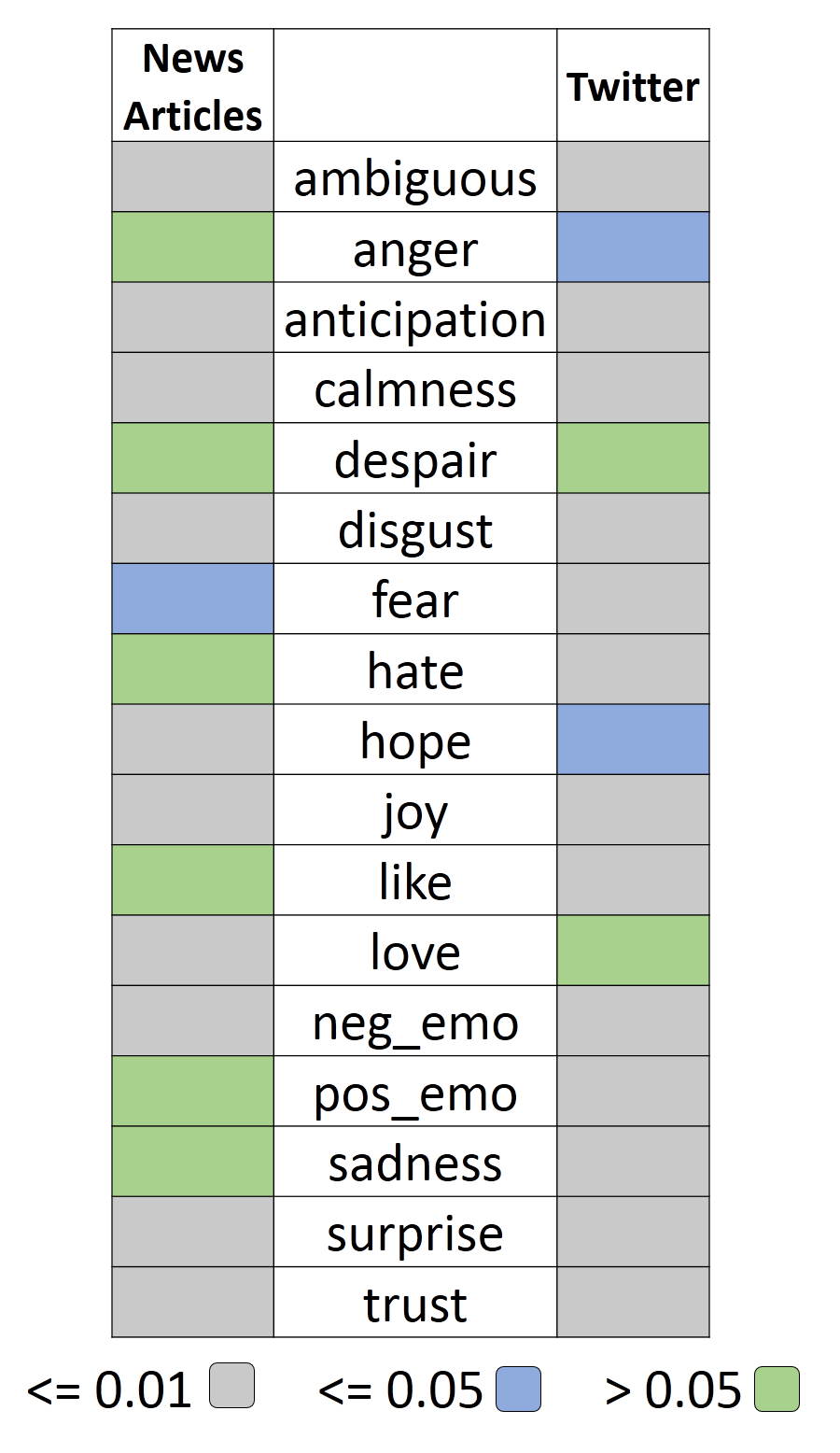}
  \caption{Statistical significant differences between false and real news on Twitter and news articles datasets using t-test.}
  \label{fig:t-test}
\end{figure}

We measure statically significant differences using the t-test on emotions across real news and false news (binary task) in the both datasets in Figure \ref{fig:t-test}. These findings provide a deeper understanding of the EIN performance. The results show that "joy", "neg\_emo", "ambiguous", "anticipation", "calmness", "disgust", "trust" and "surprise" have significant statistical differences between real and suspicious news in both datasets. Some other emotions such as "despair" and "anger" have no statistical difference in both datasets. It turns out that the results we obtain are generally consistent with the IG results in research question \textbf{RQ2}. We notice in the IG analysis that some emotions have a higher importance in one of the news sources: "sadness", "anger", and "fear" have a higher importance in Twitter than in news articles, and the opposite for "hope". We observe the same findings using the t-test.

{ \color{white}.}

\textbf{RQ4} \textit{What are the top-N emotions that discriminate false information types in both textual sources?} 

False information types are different in the way they present the news to the reader. This raises a question: what are the top employed emotions in each type of false information? In Table \ref{tab:top_emo}, we present the first three\footnote{We used SVM classifier coefficients (linear kernel) to extract the most important emotions to each classification class.} emotions that contribute mostly to the classification process to each type. This can indicate to us what are the emotion types that are used mostly in each type of false information. 

\begin{table}%
\caption{The top 3 most important emotions in each false information type.}
\label{tab:top_emo}
\begin{minipage}{\columnwidth}
\begin{center}
\begin{tabular}{ccccc}
  \toprule
  Rank & clickbait & hoax & propaganda & satire \\
  \hline
  \multicolumn{5}{c}{\textbf{News Articles}}\\
  \hline
  1 & surprise & hope & joy & disgust \\
  2 & neg\_emo & anger & fear & neg\_emo \\
  3 & like & like & calmness & pos\_emo \\
  \bottomrule
  \hline
  \multicolumn{5}{c}{\textbf{Twitter}}\\
  \hline
  1 & surprise & like & fear & pos\_emo \\
  2 & neg\_emo & disgust & hope & disgust \\
  3 & fear & anticipation & calmness & sadness \\
  
\end{tabular}
\end{center}
\centering
\end{minipage}
\end{table}%

\begin{figure*}[!htb]
  \includegraphics[width=14cm]{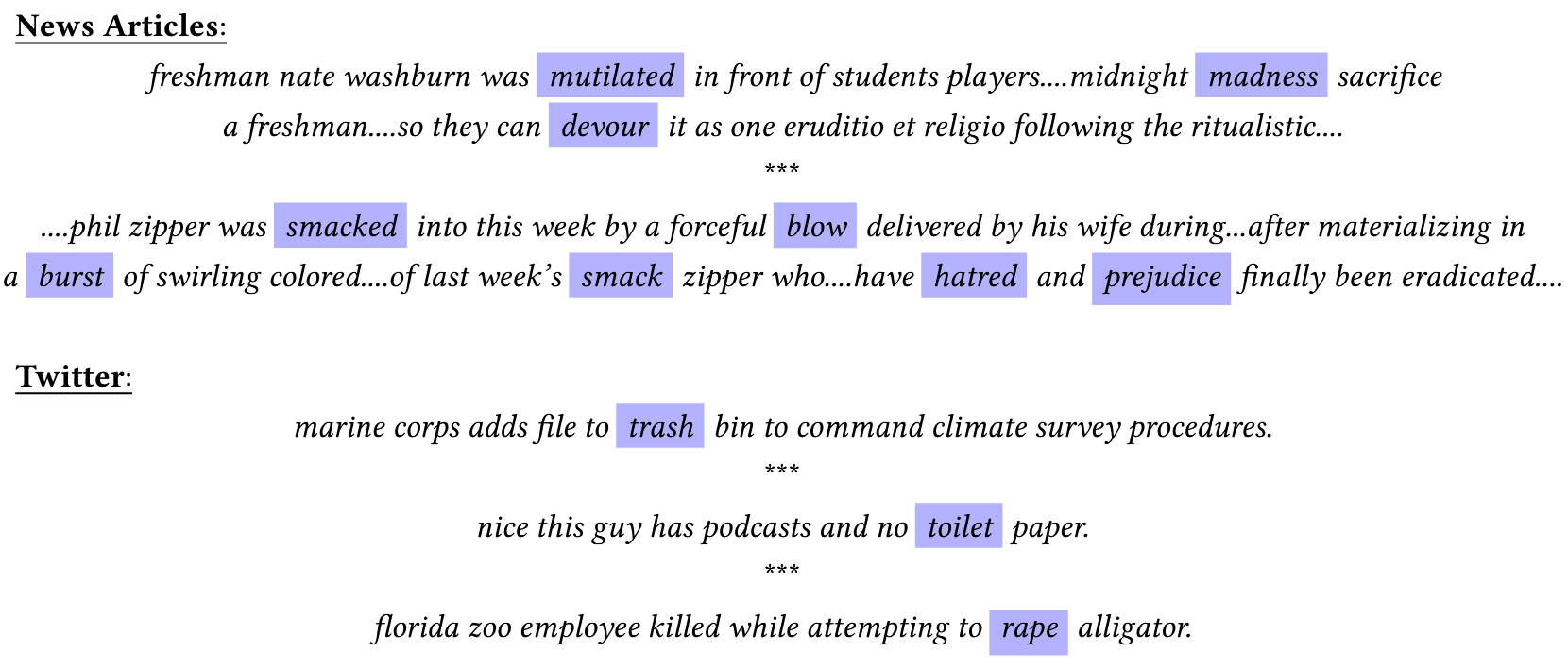}
  \caption{Examples from news articles and Twitter datasets trigger the emotion "disgust".}
  \label{fig:disgust}
\end{figure*}

Table \ref{tab:top_emo} shows that clickbaits express "surprise" and "negative emotion" at the most. This validates the definition of clickbaits as "attention redirection" by exploiting the reader and convincing him/her that there is an unexpected thing with negative emotion. The result of seeing "fear" in the top features in Twitter is interesting; one of the recent studies is presenting the hypothesis that says: \textit{curiosity is the best remedy for fear}~\cite{livio2017makes} based on psychological interpretations. Taking into account the definition of clickbaits as "attention redirection", looking at our results, we can proof this hypothesis.
Furthermore, despite the language differences in both datasets, we obtain almost the same results, which emphasize our results. For {hoaxes}, it is not simple to interpret a specific pattern of emotions in the results. We might justify it by the fact that hoaxes are written to convince the reader of the validity of a story. Therefore, the writer is trying to present the story in a normal way (truthful) similar to a real story. Therefore, the top emotions are not unique to the hoax type. But what we find from the top hoaxes emotions in both datasets is that they are generally different except the emotion "like". Despite the natural narrative way of presenting the story, the analysis shows that the writer still uses "like" to grab reader's attention smoothly. Propaganda type has clearer emotional interpretation considering its definition. We find that propaganda expresses "joy", "fear" and at the same time "calmness" in the news articles. Both "joy" and "fear" are contrary from an emotional polar perspective, where "joy" shows the extreme of the positive emotions and "fear" the extreme negative, and at the same time, "calmness" is present. The emotional shifting between the two extremes is a clear attempt of opinion manipulation from an emotional perspective. We obtain a similar emotion set from Twitter, but instead of "joy" we get "hope". Lastly, {satire} is defined as a type of parody presented in a typical format of mainstream journalism, but in a similar way to irony and sarcasm phenomena~\cite{rubin2015deception}. 
The results of the analysis show that "disgust" and "positive emotion" are present in both datasets, but we get "negative emotion" in the news articles and "sadness" in Twitter (both are placed in the negative side of emotions). 
We are interested in investigating the cause of the emotion "disgust" which appeared in the results from both datasets. We conduct a manual analysis on the text of the satire type in both datasets in order to shed some light on the possible causes. 
We notice that the satire language in the news often employs the emotion "disgust" to give a sense of humor. Figure \ref{fig:disgust} shows some examples from the news articles dataset highlighting the words that triggered the emotion "disgust".



\section{Conclusions and Future Work}
\label{CF}
In this article we have presented an emotionally-infused deep learning network that uses emotional features to identify false information in Twitter and news articles sources. We performed several experiments to investigate the effectiveness of the emotional features in identifying false information. We validated the performance of the model by comparing it to a LSTM network and other baselines.
The results on the two datasets showed that clickbaits have a simpler manipulation language where emotions help detecting them. This demonstrates that emotions play a key role in deceiving the reader. Based on this result, we investigated our model performance on a clickbaits dataset and we compared it to the state-of-the-art performance. Our model showed superior results near to 96\% F1 value.

Overall results confirmed that emotional features have boosted EIN model performance achieving better results on 3 different datasets (\textbf{RQ1}). These results emphasized the importance of emotional features in the detection of false information.  
In Twitter, false news content is deliberately sexual oriented and it uses many insulting words. Our analysis showed that emotions can help detecting false information also in Twitter. 

In the analysis section, we answered a set of questions regarding the emotions distribution in false news. We found that emotions have similar importance distribution in Twitter and news articles regardless of the differences in the used languages (\textbf{RQ2}). The analysis showed that most of the used emotions have statistical significant difference between real and false news (\textbf{RQ3}). Emotions plays a different role in each type of false information in line with its definition (\textbf{RQ4}). We found that clickbaits try to attract the attention of the reader by mainly employing the "surprise" emotion. Propagandas are manipulating the feelings of the readers by using extreme positive and negative emotions, with triggering a sense of "calmness" to confuse the readers and enforcing a feeling of confidence. Satire news instead use the "disgust" emotion to give a sense of humor. To sum up, we can say that the initial part of false news contains more emotions than the rest of document. Our approach exploit this fact for their detection.


To the best of our knowledge, this is the first work that analyzes the impact of emotions in the detection of false information considering both social media and news articles.
As a future work, the results of our approach as a clickbaits detector motivate us to develop for a clickbaits detector as a web browser extension.
Also, we will study how the emotions flow inside the articles of each kind of false information, which is worthy to be investigated as the results of this work confirmed.





%
\bibliographystyle{ACM-Reference-Format}
\bibliography{sample-base.bib}

%

\end{document}